%% file: acl2023.tex
\pdfoutput=1

\documentclass[11pt]{article}

\usepackage[]{ACL2023}

\usepackage{times}
\usepackage{latexsym}
\usepackage{tikz}
\usetikzlibrary{decorations.pathreplacing,positioning,shadows}
\usetikzlibrary{trees}
\usepackage{lipsum}
\usepackage{booktabs}
\usepackage{multirow}
\usepackage{listings} 
\usepackage{xcolor} 
\usepackage{pifont}
\usepackage[T1]{fontenc}
\usepackage{color,soul}
\usepackage[utf8]{inputenc}
\usepackage{enumitem}
\usepackage{inconsolata}
\usepackage{bbm}

\usepackage{tikz}
\tikzset{
  font={\fontsize{6.5pt}{1}\selectfont}}
\usetikzlibrary{decorations.pathreplacing,positioning,shadows}
\usetikzlibrary{trees}
\usepackage{lipsum}
\usepackage{amsfonts}

\usepackage[T1]{fontenc}

\usepackage[utf8]{inputenc}

\usepackage{microtype}

\usepackage{inconsolata}

\usepackage{tcolorbox}
\tcbuselibrary{breakable}
\definecolor{yellow}{HTML}{F6BD60}
\definecolor{white}{HTML}{F7EDE2}
\definecolor{pink}{HTML}{F5CAC3}
\definecolor{tale}{HTML}{84A59D}

\definecolor{cadmiumgreen}{rgb}{0.0, 0.42, 0.24}
\definecolor{celestialblue}{rgb}{0.29, 0.59, 0.82}

\newcommand{\myblue}[1]{\textcolor{celestialblue}{#1}}
\title{
Large Language Models Fall Short: Understanding Complex Relationships in Detective Narratives
}

\author{Runcong Zhao$^{1*}$, Qinglin Zhu$^{1*}$,  Hainiu Xu$^1$, Jiazheng Li$^1$, Yuxiang Zhou$^1$\\ 
\textbf{Yulan He$^{1,2,3}$, Lin Gui$^1$}\\
  $^1$King's College London, $^2$University of Warwick, $^3$The Alan Turing Institute\\
  \texttt{\{runcong.zhao, qinglin.1.zhu, hainiu.xu, jiazheng.li, yuxiang.zhou\}@kcl.ac.uk} \\ \texttt{\{yulan.he, lin.1.gui\}@kcl.ac.uk} }

\begin{document}
\maketitle
\def\thefootnote{*}\footnotetext{Equal contribution.}\def\thefootnote{\arabic{footnote}}
\begin{abstract}
Existing datasets for narrative understanding
~often fail to represent the complexity and uncertainty of relationships in real-life social scenarios. To address this gap, we introduce a new benchmark, \emph{Conan}, designed for extracting and analysing intricate character relation graphs from detective narratives. 
Specifically, we designed hierarchical relationship categories and manually extracted and annotated role-oriented relationships from the perspectives of various characters, incorporating both public relationships known to most characters and secret ones known to only a few. 
Our experiments with advanced Large Language Models (LLMs) like GPT-3.5, GPT-4, and Llama2 reveal their limitations in inferencing complex relationships and handling longer narratives. 
The combination of the \emph{Conan} dataset and our pipeline strategy is geared towards understanding the ability of LLMs to comprehend nuanced relational dynamics in narrative contexts.
\end{abstract}

\input{Sections/introduction_v1}

\section{Task Definition}

\begin{figure}[th!]
    \centering
    \includegraphics[width=\linewidth]{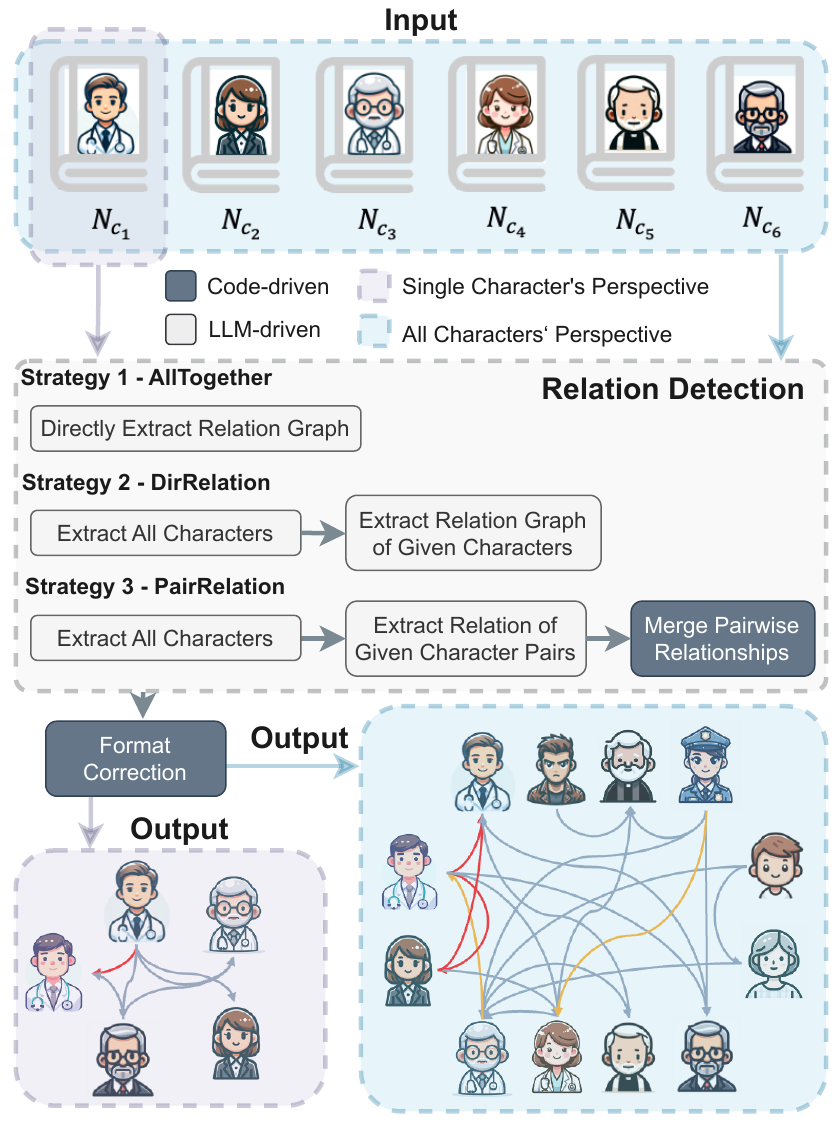}
    \caption{Input-Output Format and Benchmark Relation Detection Strategies. The input narrative consists of $k$ background stories $N_{c_i}$ that are uniquely created from the perspective of the character $c_i$. Our objective is to extract all characters from the given story, including those beyond the initial $k$ characters,  subsequently detect the relationships among all the extracted characters, even when they involve false or multiple identities, and finally uncover conflicting relationships in order to deduce the genuine nature of these relationships. }
    \label{fig:setting}
\end{figure}


As shown in Figure~\ref{fig:setting}, the input narrative $N$ consists of background stories for $k$ characters, represented as $c_i \in C_k = \{c_1, c_2, ..., c_k\}$. The background story of each character $N_{c_i}$ is crafted solely from the perspective of that particular character $c_i$, with all relationships and events in the story framed based on $c_i$'s perception. Unlike the conventional narrative structure that revolves around a single protagonist, in our setup, the complete narrative $N$ emerges as a collaborative novel,  composed by intertwining the perspectives of these $k$ characters, i.e., $N = \cup_{c_i \in C_k} N_{c_i}$.

As depicted in Figure~\ref{fig:relationships}, for the same set of character relationships, there can be instances where some characters remain completely unaware of these connections, or where the perceptions of different characters are in direct contradiction. Moreover, the narrative may introduce additional characters beyond these $k$ individuals. Some of these extra characters might play peripheral roles, while others could hold pivotal significance, such as the victim. Therefore, we define all characters that appeared in the story as $C = \{c_1, c_2, ..., c_K\}$, where $C_k \subseteq C$. Hence, by defining $R_C$ as the relationships among this group of characters $C$, we aim to perform the following three sub-tasks:
\begin{enumerate}
\item \textbf{Character Extraction}. 
Identify all characters in the given story, which can be $N_{c_i}$ or $N$. 

\item \textbf{Entity Linking}. Recognise character relationships from the perspective of a specific character $c_i$. 
This equals $R_C|N_{c_i}$ formally.

    
\item \textbf{Relation Deduction}. Infer the actual relationships between characters by considering the collection of all character-centric narratives. Formally, this corresponds to $R_C|N$. 
\end{enumerate}

To understand the relationships of characters within detective narratives, the challenges mainly manifested in three types, corresponding to the aforementioned three sub-tasks: First, in detective narratives, a character might have different identities or aliases, and these identities may surface at different points in the story. 
Secondly, LLMs struggle with basic inference based on existing information due to a phenomenon known as reversal curse \citep{reversal-curse2023}. For example, when we know that \textit{A} is \textit{B}'s father, it does not necessarily imply that \textit{B} is \textit{A}'s child according to these models. 
Finally, the third challenge involves drawing inferences from the perspectives of multiple characters, which sometimes yield conflicting or inconsistent information. For instance, from \textit{A}'s perspective, \textit{B} might be regarded as his mother's friend. However, from \textit{B}'s perspective, he has been the lover of \textit{A}'s mother and is actually \textit{A}'s biological father. 

\section{Dataset Construction}

To investigate the capabilities of LLMs in comprehending detective narratives, we developed the first open benchmark dataset for this task. The construction process includes data collection and processing, and annotation of characters and their relationships. 

\subsection{Data Collection and Processing}
\paragraph{Collection} 
We discovered that the background narratives crafted for murder mystery games are particularly suitable for several reasons. Firstly, they feature complex character relationships that often pose challenges for LLMs. Secondly, these narratives offer more realistic scenarios compared to existing synthetic benchmarks. Lastly, while these stories are considerably longer than current benchmarks, they still maintain a level of consciousness suitable for human reading
. This equilibrium ensures that the workload for human annotators, in terms of reading and comprehending these narratives, remains manageable. 

\paragraph{Filtering} We selected 100 high quality narratives from a pool of $2,135$, following the filter process described in Appendix \ref{sec:data-selection}. Then we employed Adobe to extract text from scanned narratives and leveraged the capabilities of the ChatGPT model GPT-3.5 to rewrite and refine the text. 

\begin{figure*}[th!]
    \centering
    \includegraphics[width=0.7\linewidth]{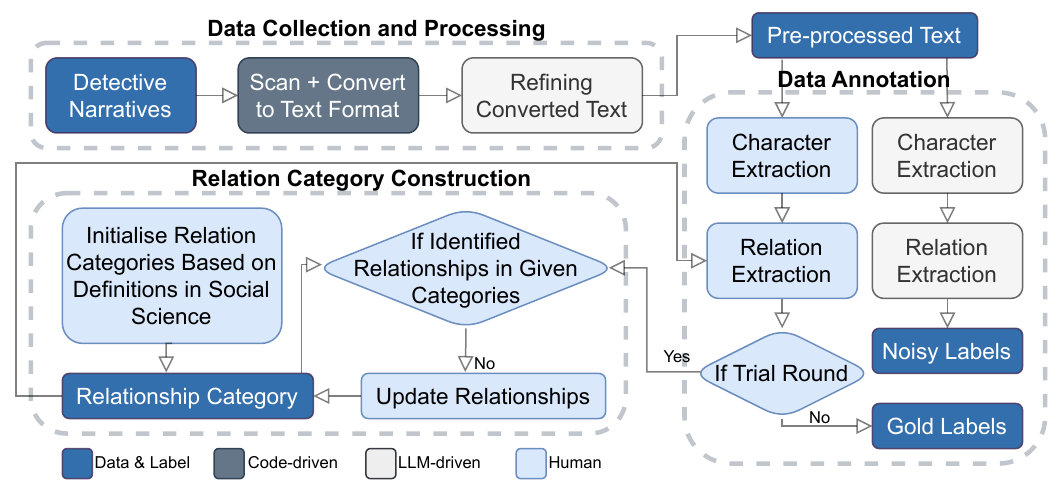}
    \caption{Dataset Construction. }
    \label{fig:data-process}
\end{figure*}

\subsection{Data Annotation and Evaluation}
\paragraph{Relation Category Construction} 
We started with relationship scheme extraction. Based on a relationship scheme of 53 casual relationships \cite{berscheid1994interpersonal, devito2019interpersonal}, we carried out a trial run using 10 detective novels to expand upon these relationships. The resulting scheme contains 5 primary relationship categories, namely \textit{Romantic}, \textit{Family}, \textit{Allies}, \textit{Adversarial}, and \textit{Business}, which consist of 163 fine-grained relationships denoted as $\mathtt{R}$. 

During this process, we found that the boundaries between some categories were not clearly defined, such as \emph{``mother''}, \emph{``mother-in-law''}, and \emph{``adoptive mother''}. In contrast, mistaking \emph{``mother''} for \emph{``stranger''} is a more significant error than confusing these three specific relationships. 
To mitigate such ambiguities, we consolidated similar categories into broader ones until no further merging was feasible. For instance, we merged \emph{``lover''}, \emph{``boyfriend''}, and \emph{``girlfriend''} into a single category: \emph{``romantic relationships''}. Consequently, we established a hierarchical structure of relationship categories, comprising 5 top-level categories, 54 intermediate categories, and 163 detailed categories. Such a well-defined, detective-oriented relationship scheme helps to reduce the potential subjectivity during the annotation process. 

Evaluating free-form relationships generated by LLMs can be extremely challenging, requiring either human evaluation or using LLMs for auto-evaluation. However, human evaluation is costly and subjective, while evaluation by LLMs has been shown to have notable disparities compared to human judgement. Therefore, it is better to annotate the relationships between characters, providing a consistent basis for 
evaluating this dataset.

\paragraph{Labelling}
We recruited four annotators, all of whom were fans of detective narratives, and conducted training sessions for them. Our complete annotation pipeline involves three tasks: (1). \emph{Character Extraction}: Annotators read given detective novels to identify all characters appearing in the narrative. (2). \emph{Entity Linking}: annotators closely examine the story from the perspective of a single character, and extract both explicit and implicit relationships. These relationships are structured as triplets in the format of $\big(c_i, c_j, r_{i,j})$, where $r_{i,j}$ signifies the relationship between characters $c_i$ and $c_j$. We specifically account for scenarios where $i = j$ to handle the common detective novel plot where a single character has multiple identities (refer to Appendix~\ref{appendix:relationship_schemes} for details). (3). \emph{Conflict Detection and Relationship Refinement}: As our dataset comprises narratives from multiple character perspectives, annotators are tasked with considering all available information to form a unified relationship graph from the $k$ individual detective narratives. This step often involves resolving conflicting information that arises from the imperfect knowledge shared among characters and refining the final relationships through inference.

Consequently, for each story, our team of experts would generate $k$ distinct relationship graphs at the individual character level, and one consolidated graph that merges these individual perspectives. 

\paragraph{Inter-annotator Agreement}
Following the agreement measure for triplets in previous works \citep{semeval2007-girju, drug-effects2012}, we used F1-score as a criterion to measure inter-annotator agreement (IAA). We selected one challenging narrative and asked three annotators to annotate it following the annotation guidelines. This step served the purpose of evaluating the inter-annotator agreement and ensuring the quality of the annotations. The detailed annotation guideline and the calculation of the inter-annotator agreement score are in Appendix~\ref{appendix:annotation_guideline}. 
We calculate IAA for three steps in our task: 1). identifying characters, $IAA(c_i)$; 2). determining if there are relationships between a given character pair, $IAA(c_i, c_j)$; and 3). classifying the relationships between two characters when there are relationships between them, $IAA(c_i, c_j, r_k)$. $IAA(c_i, c_j, r_k)_{1}$ assesses whether annotators agree on at least one relationship for each character pair. $IAA(c_i, c_j, r_k)_{all}$ measures agreement on all relationship triplets. As shown in Table~\ref{tab:agreement-score}, annotators demonstrate high agreement in both character and relationship extraction. However, agreement decreases as the task becomes more subjective. For instance, identifying characters present in the narrative is more straightforward, but \emph{``father's friend of x''} might be labelled as \emph{``acquaintance of x''} by one annotator and as unrelated by another.

\begin{table}[h]
\centering
\resizebox{\linewidth}{!}{
\begin{tabular}{ccccc}
\toprule
                               &        Character                     & \multicolumn{3}{c}{Relation}           \\ \cmidrule(lr){2-2} \cmidrule(lr){3-5}
\multirow{-2}{*}{Annotator}    & $c_i$ & $(c_i, c_j)$ & $(c_i, c_j, r_k)_{1}$ & $(c_i, c_j, r_k)_{all}$ \\ \midrule
{1 \& 2} & 0.978                       & 0.894    & 0.962        & 0.873 \\
{1 \& 3} & 0.978                       & 0.800    & 0.916        & 0.738 \\
{2 \& 3} & 0.966                       & 0.756    & 0.907        & 0.736 \\ \midrule
Average                        & 0.974                       & 0.817    & 0.928        & 0.782 \\
\bottomrule
\end{tabular}}
\caption{Inter-annotator Agreement Score. } 
\label{tab:agreement-score}                       
\end{table}

\subsection{Data Statistics}

The detailed statistics are shown in Table~\ref{tab:dataset_statistics}. We collected a total of 2,135 narratives, but the majority of them exhibited low quality. After manual selection, we identified 100 high-quality narratives. We first generated annotations automatically using GPT-4, categorised under the column \emph{All} in Table~\ref{tab:dataset_statistics}. Our dataset has an average of 27,695 tokens per narrative, making it significantly longer and more complex than earlier datasets, particularly considering its detective-themed content.

We also recruited human experts to annotate 25 narratives, resulting in a total of 8,254 annotations. They were compensated at an hourly rate of \$31.92, with each narrative estimated to take about 10 hours to complete. Due to quality concerns, we removed one annotated narrative from our final dataset as it ambiguously described the relationships between characters, making the annotation highly subjective. We ended up with 24 annotated narratives encompassing 7,951 relationships.

The original narrative is in Chinese, but we also provide an English-translated version and conduct experiments on it, detailed in Appendix~\ref{sec:translated version}.

\section{Experiments}
\subsection{Baselines}
We conducted our experiments using OpenAI's models and Llama2-chat. We accessed GPT-3.5 and GPT-4 through the Azure API with model version gpt-35-turbo-16k 0613(Default) and gpt-4-turbo 1106-Preview. The experiments were carried out with the default parameters of the interface, between October and December 2023. For Llama2, we used the HuggingFace model Llama-2-70b-chat-hf. Inferences on this model were run directly using greedy decoding at a temperature setting of 0. 

\begin{table}[h]
\centering
\resizebox{0.9\columnwidth}{!}{
\begin{tabular}{@{}lrr@{}}
    \toprule
    \textbf{Dataset}  & \textbf{All} & \textbf{Human} \\
    \midrule
     \#Narratives & 100 & 24 \\
     \#Background stories & 640 & 149\\
    Avg. \#Character per narrative  & 18.72 & 18.84 \\ 
    \ \ \ \ \ w/ narrative    & 7.40   & 7.21 \\
    \ \ \ \ \ w/o narrative    &  11.32  &  11.63\\
    \#Relationships & 27,444 & 7,951 \\
    Avg. \#Token per character story   & 4,327 & 4,539 \\
    Avg. \#Token per narrative & 27,695 & 28,182\\
    \bottomrule
\end{tabular}
}
\caption{Dataset Statistics.
}

\label{tab:dataset_statistics}
\end{table}


Given that the length of some narratives exceeds the input limit of LLMs, we segment the original narrative input into multiple parts. The relation graph is initially extracted from the first segment. Subsequently, this pre-established relation graph, coupled with the ensuing narrative segment, is used as input to instruct the model to update and refine the relationship graph. This iterative process allows for comprehensive relationship mapping despite the constraints of LLM input limitations.

We evaluate relation detection using three strategies: 
First, \emph{AllTogether}, where we ask the model to directly output a relationship graph, including characters and their relationships. Second, \emph{DirRelation}, 
to distinguish and minimise the influence of errors occurring in two separate stages, character extraction, and relationship extraction, we extract the characters from the narrative first, then utilise it alongside the narrative script to generate the relationship network.
Third, \emph{PairRelation}, where we initially extract characters, inquire about the relationship of each character pair, and finally, aggregate the results, merging them into a comprehensive relationship map. This strategy targets LLM's problem of 
ignoring relationships in narratives, especially with long narratives. 

The total running costs for using GPT-3.5 and GPT-4 in our experiments are approximately \$400 and \$2000, respectively. 
In addition, the running time for Llama2 in our experiments totalled 490 hours, utilising two 80G A100 graphics cards.

\subsection{Corruption Rate}
For generative language models, there is always a possibility that the output format may not follow the given instructions. Even though these models, 
are trained to follow specific formats like JSON, complex tasks such as \emph{Conan} still demonstrate instances where models fail to comply with the provided format guidelines. 
Therefore, we classify cases failing to produce the desired format as corrupted cases. We calculate the corruption rate for various models and strategies as $\frac{n_c}{n_{all}}$
, where $n_c$ is the number of corrupted relationships and $n_{all}$ is the number of total generated relationships.
For each model, we add a post-processing step to remap the relationships not in the specified categories into desired categories. However, some outputs remain corrupted even after recategorising. 

\begin{table}[ht!]
\centering
\resizebox{\linewidth}{!}{
\begin{tabular}{ccccccc}
\toprule
\multirow{2}{*}{Strategy}                    &       \multicolumn{2}{c}{Llama2} & \multicolumn{2}{c}{GPT-3.5} & \multicolumn{2}{c}{GPT-4} \\ \cmidrule(lr){2-3} \cmidrule(lr){4-5} \cmidrule(lr){6-7}
& before & after & before & after & before & after \\ \midrule
\multicolumn{7}{c}{\textit{Corruption Rate}} \\ 
\midrule
AllTogether         &    0.445 &    0.316  & 0.310 & 0.032 & 0.143 & \textbf{0.006}   \\
DirRelation         &   0.448 &   0.286   & 0.266    & 0.022 & 0.164   & \textbf{0.009}  \\
PairRelation         &   0.563   &   0.336  &  0.160 &  \textbf{0.021} & 0.180  & 0.032   \\

\midrule
\multicolumn{7}{c}{\textit{F1-score}} \\ 
\midrule

AllTogether         &  0.030   & \textbf{0.035}    &  0.028  & \textbf{0.031 }   &  \textbf{0.125}  & 0.119   \\
DirRelation         &   0.050  & \textbf{0.057}    &  0.052  & \textbf{0.053}  & \textbf{0.110} &  0.103  \\
PairRelation        &   0.020  & \textbf{0.021}    &  0.025  & \textbf{0.025 }   &  \textbf{0.027}  &  0.026 \\

\bottomrule
\end{tabular} }
\caption{Comparison of corruption rate and F1-score before and after self-correction. The corruption rate is the percentage of output relationships that fail to comply with the provided format guidelines before and after self-correction. 
} 
\label{tab:evaluation-corruption-rate}                       
\end{table}

As indicated in Table~\ref{tab:evaluation-corruption-rate}, there was a significant reduction in the corruption rate following post-processing. However, our primary focus is the F1-score after self-correction. This step proved beneficial for Llama2 and GPT-3.5, but counterproductive for the best-performing model, GPT-4. Consequently, we chose to implement the self-correction step only for Llama2 and GPT-3.5 in our subsequent experiments.

\subsection{Character Extraction} 
Besides the F1-score of character extraction for three baseline relation detection strategies, we also report the results that we ask LLMs to directly extract characters from the given narrative, which were used in DirRelation and PairRelation. We denote it as DirCharacter in Table~\ref{tab:evaluation-character}. 
\begin{table}[ht!]
\centering
\resizebox{\linewidth}{!}{
\begin{tabular}{cccccccccc}
\toprule
\multirow{2}{*}{Strategy} & \multicolumn{3}{c}{Llama2} & \multicolumn{3}{c}{GPT-3.5} & \multicolumn{3}{c}{GPT-4} \\ \cmidrule(lr){2-4} \cmidrule(lr){5-7} \cmidrule(lr){8-10}
  & Pre      & Rec     & F1     & Pre     & Rec    & F1    & Pre    & Rec    & F1   \\ 
  \midrule
  \multicolumn{10}{c}{\textit{Information from Single Character's Perspective}}  \\ \midrule
DirCharacter  & 0.636 & 0.605 & 0.620 & 0.664 & 0.697 & 0.680 & 0.613 & 0.782 & 0.687 \\
AllTogether   & 0.665 & 0.489 & 0.564 & 0.229 & 0.373 & 0.283 & 0.755 & 0.789 & \textbf{0.772} \\
DirRelation   & 0.725 & 0.540 & 0.619 & 0.644 & 0.678 & 0.660 & 0.658 & 0.768 & 0.709 \\
PairRelation  & 0.575 & 0.615 & 0.594 & 0.643 & 0.714 & 0.677 & 0.612 & 0.780 & 0.686 \\
\midrule
\multicolumn{10}{c}{\textit{Information from All Characters' Perspectives}} \\ 
\midrule
DirCharacter  & 0.449 & 0.703 & 0.548 & 0.522 & 0.836 & \textbf{0.643} & 0.444 & 0.870 & 0.588 \\
AllTogether   & 0.671 & 0.261 & 0.376 & 0.240 & 0.238 & 0.239 & 0.736 & 0.455 & 0.562 \\
DirRelation   & 0.730 & 0.332 & 0.457 & 0.502 & 0.517 & 0.509 & 0.681 & 0.442 & 0.536 \\
PairRelation  & 0.371 & 0.598 & 0.458 & 0.478 & 0.844 & 0.610 & 0.430 & 0.869 & 0.575 \\

\bottomrule
\end{tabular}}
\caption{Character extraction results.} 
\label{tab:evaluation-character}
\end{table}
We can see that GPT-4 performs the best when extracting information from a single character’s perspective. However, its performance drops significantly when extracting information based on all characters' perspectives, a more pronounced drop than observed with Llama2 and GPT-3.5. Upon comparing extracted characters based on all characters' information using different LLMs, we noted that GPT-4 tends to extract more details, averaging 31.92 characters per narrative. In contrast, Llama2 extracted 25.54 characters, and GPT-3.5 extracted 26.08 characters, which are significantly less than the GPT-4's extraction. Consequently, this results demonstrate higher recall and precision for GPT-4 with shorter inputs, in comparison to the aforementioned models. Yet, for longer narratives, GPT-4's precision suffers due to issues like character duplication (e.g., ``Costa'', ``Head Nurse Costa'', and ``Sylvia Costa'' all referring to the same character) and misclassification of entities such as organisation names.

\subsection{Relation Extraction} 
After removing the corrupted cases, we calculate the F1-score based on the derived relationship triples. These triples consist of character $i$, character $j$, and the relationships between them, represented as $(c_i, c_j, r_{i,j})$. Here we define precision as $\frac{n_{p}}{n_g}$, where $n_{g}$ is the number of generated triples after removing corrupted ones, $n_{p}$ is the number of correct relationship triples among generated ones; and recall as $\frac{n_{r}}{n_l}$, where $n_{l}$ is the number of labelled triples, $n_{t}$ is the number of matched triples among labelled ones.


\begin{table}[ht!]
\centering
\resizebox{\linewidth}{!}{
\begin{tabular}{cccccccccc}
\toprule
\multirow{2}{*}{Strategy} & \multicolumn{3}{c}{Llama2} & \multicolumn{3}{c}{GPT-3.5} & \multicolumn{3}{c}{GPT-4} \\ \cmidrule(lr){2-4} \cmidrule(lr){5-7} \cmidrule(lr){8-10}
  & Pre      & Rec     & F1     & Pre     & Rec    & F1    & Pre    & Rec    & F1   \\ \midrule

\multicolumn{10}{c}{\textit{Information from Single Character's Perspective}}  \\ \midrule
AllTogether  & 0.129 & 0.054 & 0.076 & 0.056 & 0.041 & 0.047 & 0.283 & 0.269 & \textbf{0.276}\\
DirRelation  & 0.160 & 0.085 & 0.111 & 0.119 & 0.076 & 0.093 & 0.219 & 0.267 & 0.240 \\
PairRelation & 0.025 & 0.089 & 0.039 & 0.029 & 0.099 & 0.045 & 0.047 & 0.258 & 0.080  \\ 
\midrule
\multicolumn{10}{c}{\textit{Information from All Characters' Perspectives}}  \\ \midrule
AllTogether &  0.121 & 0.020 & 0.035 & 0.064 & 0.020 & 0.031 & 0.267 & 0.082 & \textbf{0.125}\\
DirRelation &  0.203 & 0.033 & 0.057 & 0.092 & 0.037 & 0.053 & 0.202 & 0.075 & 0.110 \\
PairRelation & 0.012 & 0.092 & 0.021 & 0.014 & 0.132 & 0.025 & 0.014 & 0.238 & 0.027\\ 
\bottomrule
\end{tabular}}
\caption{F1-score of Relation Extraction.} 
\label{tab:evaluation-f1}
\end{table}

As illustrated in Table~\ref{tab:evaluation-f1}, it is evident that both Llama2 and GPT-3.5 struggle significantly to extract relationships from long narratives. In comparison, GPT-4 demonstrates considerably better performance, although there remains a substantial gap compared to human understanding. As anticipated, extracting relationships based on information from all characters' perspectives is a more challenging task compared to extracting relationships based on information from a single character's perspective. This increased difficulty arises due to two factors: 

\paragraph{More Complicated Information} Narratives from a single character's perspective typically don't contain self-contradictory content. This is because the information is based on that character's perception, which often tends to be consistent and self-explanatory. Therefore, LLMs can just extract what is stated in the text.
In contrast, information from all characters includes secrets, misunderstandings, lies generated for self-interest or specific goals, and even delusions caused by illnesses. Deriving accurate character relationships from these potentially repetitive or contradictory pieces of information is naturally more complex. LLMs must navigate through various narrative layers, distinguishing between truth, deception, and perception to accurately infer relationships. 

To validate our assumption, we also calculate the accuracy of the relationships that are inconsistent across different characters' perspectives, which we identify as secret or inferred relations, as illustrated in Figure~\ref{fig:relationships}. We can see that the accuracy of those complicated relationships is lower compared to all labelled relationships in Table~\ref{tab:evaluation-f1-factors}. 

\begin{table}[h]
\centering
\resizebox{\linewidth}{!}{
\begin{tabular}{ccccccc}
\toprule
\multirow{2}{*}{Strategy}                    &       \multicolumn{2}{c}{Llama2} & \multicolumn{2}{c}{GPT-3.5} & \multicolumn{2}{c}{GPT-4} \\ \cmidrule(lr){2-3} \cmidrule(lr){4-5} \cmidrule(lr){6-7}
& all & complex & all & complex & all & complex \\ \midrule
AllTogether  & 0.020 & 0.012 & 0.018 & 0.017 & 0.082 & 0.072 \\
DirRelation  & 0.343 & 0.014 & 0.028 & 0.028 & 0.101 & 0.090 \\
PairRelation & 0.093 & 0.047 & 0.091 & 0.081 & 0.229 & 0.193 \\
\bottomrule
\end{tabular}}
\caption{Comparison with Complex Relationships. The accuracy of all relationships and only the complex relationships. 
} 
\label{tab:evaluation-f1-factors}
\end{table}

\paragraph{Longer Narrative} 
One reason we suspect for the low F1 scores is the ``lost in the middle'' phenomenon observed in long narratives, which has also been noticed earlier \citep{liu2023lost-in-the-middle}, where LLMs tend to put more attention on the beginning and the end of long inputs, often ignoring information in the middle. Consequently, additional information can enhance judgement for humans, however, when all characters' information is combined as input for LLMs, this extra data does not lead to improved results.

\begin{figure}[th!]
    \centering
    \includegraphics[width=\linewidth]{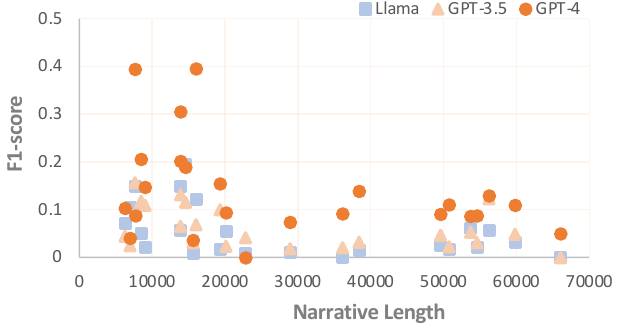}
    \caption{F1-score against the length of given narrative.}
    \label{fig:data-process}
\end{figure}

To validate our assumption, we plotted the F1-scores for each narrative against their respective lengths. The results suggest that longer narratives tend to yield lower F1-scores. However, shorter narratives do not guarantee higher F1-scores, as the results also heavily depend on the complexity of each narrative. Therefore, narratives with short lengths but complicated relationships can 
exhibit lower F1-scores due to their inherent difficulty.

\subsection{Ablation Studies}
\paragraph{Impact of Character Extraction} To investigate the impact of character list quality on relationship extraction outcomes, we assessed the performance using both a gold standard and a model-generated noisy character list, as detailed in Table~\ref{tab:evaluation-f1-compare}. Results show a significant performance increase with the provided characters.
Table~\ref{tab:evaluation-f1} reveals that while both Llama2 and GPT-3.5 showed improved results with DirRelation, GPT-4 performed better using AllTogether. Given GPT-4 did not do well in character extraction, its performance sets a ceiling for the efficacy of DirRelation and PairRelation, as their outcomes hinge on the quality of separately extracted characters. 

\begin{table}[ht!]
\centering
\resizebox{0.95\linewidth}{!}{
\begin{tabular}{ccccccc}
\toprule
\multirow{2}{*}{Strategy} & \multicolumn{2}{c}{Llama2} & \multicolumn{2}{c}{GPT-3.5} & \multicolumn{2}{c}{GPT-4} \\ \cmidrule(lr){2-3} \cmidrule(lr){4-5} \cmidrule(lr){6-7}
& gold      & noisy     & gold      & noisy     & gold      & noisy      \\ \midrule
\multicolumn{7}{c}{\textit{Information from Single Character’s Perspective}}  \\ \midrule
DirRelation     & 0.124 & 0.111 & 0.123 & 0.093 & \textbf{0.315} & 0.240 \\
PairRelation    & 0.071 & 0.039 & 0.083 & 0.045 & 0.167 & 0.080 \\
\midrule
\multicolumn{7}{c}{\textit{Information from All Characters’ Perspectives}}  \\ \midrule
DirRelation      & 0.053 & 0.057 & 0.064 & 0.053 & \textbf{0.171} & 0.110 \\
PairRelation     & 0.046 & 0.021 & 0.055 & 0.025& 0.113 & 0.027 \\

\bottomrule
\end{tabular}}
\caption{Impact Assessment of Character Extraction. We evaluated the impact on relationship extraction's F1-score using both the gold standard character list and the model-generated noisy character list.
} 
\label{tab:evaluation-f1-compare}
\end{table}

\paragraph{Impact of Strategies}
To investigate the impact of various relation detection strategies, we compared the approaches previously discussed. We noticed that directly asking models to extract all relationships resulted in low recall. We also found that when inquiring complex relationships between the given two characters, they perform better. Therefore, a straightforward solution was to inquire about the relationships between each pair of characters, which we termed as the PairRelation strategy. While this approach did increase recall, it also significantly amplified hallucinations. For instance, when the relationship between characters $a$ and $b$ wasn't explicitly mentioned, the model often fabricated one. 
Additionally, this method is more costly and time-consuming: while originally we needed to calculate it at a complexity of $O(k)$, it now escalates to $O(k^3)$, where $k$ is the number of characters. This translates to a hundredfold increase in cost for narratives with 10 characters. 

Consequently, we infer that PairRelation is the least effective strategy. For GPT-4, when a reliable list of characters is available, DirRelation should be employed. In scenarios where a high-quality character list cannot be ensured, the AllTogether approach is preferable. For Llama2 and GPT-3.5, DirRelation is always the better strategy. 

\section{Related Works}
\paragraph{Character Extraction}
Named Entity Recognition (NER) is a longstanding challenge in NLP \citep{ner-wu2020, ner-aly2021}. It involves locating and classifying named entities present in unstructured text. Character extraction is a specific area within NER, focusing on identifying characters involved in a given narrative \citep{speaker-identification2013, bamman-etal-2020-annotated, sang-etal-2022-tvshowguess}. In this process, models often confuse the targeted entities (characters) with other types of entities, such as organisations, items, locations, and so on. Another challenge is to merge all expressions in a given text that refer to the same entity, which is known as coreference resolution \citep{coreference-tvshow-chen2017, coreference-seq2seq-mou2023}. This task becomes even more challenging when dealing with long-distance mentions \citep{character-relationships-literary2015, phrase-detectives2023}. In detective narratives, misidentified characters, entwined relationships, and secrets hidden by characters can further complicate this task \citep{narrativeplay2023}.

\paragraph{Character-centric Information Extraction}
Depending on various motivations, character-centric information extraction such as identifying motivations and emotional reactions \citep{rashkin-etal-2018, emotional-relationships2019}, roles \citep{heroes-stammbach-etal-2022}, appearance \citep{narrativeplay2023}, or personalities \citep{personality-profiling2015, fictional-characters2023} can be conducted from the perspectives of these identified characters.
Relation extraction plays a key role in character-centric information extraction \citep{social-networks2009, extraction-character-networks2019}. Previous research exploring the narrative comprehension and inferential abilities of LLMs has typically relied on question answering \cite{yang2019friendsQA, fairytaleqa-xu-etal-2022, gandhi2023bigtom}, which may include questions about relationships but do not cover all aspects of character relationships. Alternatively, some research has been conducted on extracting relationships directly from sentences \cite{you2020relation-trajectories, mallace2020relation-clustering}, typically focusing on verbs, such as \emph{say}, \emph{smile}, \emph{look}.
Relation extraction is frequently integrated with other character-centric tasks like dialogue generation \citep{chen2023harrypotter}, summarisation \citep{brahman-etal-2021-characters-tell}, and tracking the evolution of relationships between characters over time \citep{dynamic-fictional-relationships2016}.

\section{Conclusion and Future Directions}
This research introduces a task for LLMs to comprehend complex character relationships in detective narratives, utilising our newly created \emph{Conan} dataset. This dataset highlights the current challenges for LLMs and aims to improve their ability in narrative contexts.
\paragraph{Challenges}
(1). \emph{Enhancing Inferential Abilities of LLMs.} Our research reveals that LLMs face challenges in deciphering complex relationships, particularly when the input narrative contains conflicting information. To address this, recent advancements \citep{chain-of-thoughts-wei2022, tree-of-thoughts-yao2023, promptagent-wang2024} can be leveraged to augment the inferential capabilities of LLMs. 
(2). \emph{Optimising Key Information Management.} We observed that LLMs struggle to pinpoint key information in lengthy inputs. Therefore, employing methods like cosine similarity \citep{generative-agent2023} or Retrieval-Augmented Generation (RAG) \footnote{\url{https://github.com/weaviate/Verba}} to assist LLMs in managing, retrieving, and focusing on the most relevant or crucial information could be beneficial. 

\paragraph{Applications}
(1). \emph{Enhancing Narrative Understanding.} Our work can be used to analyse complex narratives in literature, films, and video games. It helps in understanding character dynamics and plot development. 
(2). \emph{Interactive Agents.} AI-driven agents are widely used in many sectors, including chatbots that support people's work and emotional requirements \citep{qian2023communicative, characterchat-tu2023}; interactive game development \citep{npc2023, narrativeplay2023}; digital life simulation \citep{digital-life2023}. Understanding the relationships between characters in user inputs can enhance conversation quality, making these systems more empathetic and context-aware. 
(3). \emph{Theory of Mind.} Our dataset, built on various characters' perspectives, naturally includes insights like how characters perceive relationships among others and how they think others view them \citep{premack1978does}. This makes it suitable for theory of mind tasks \citep{wimmer1983beliefs, onishi200515}.

\section{Limitations}
\paragraph{Subjectivity Inherited from Annotators}
Despite our efforts to standardise annotation guidelines through trial rounds, interpretations of relationships in \emph{Conan} may still carry a degree of subjectivity. Annotators’ perceptions can influence the representation of character dynamics, leading to variability in relationship categorisation.

\paragraph{Limited Annotation Scope} The high cost of manual annotation constrained our ability to label all 100 narratives with human annotators. To address this limitation, we employed the highest-performing model and strategy for annotation, which may not fully capture the relationships in the given narratives.

\paragraph{Restricted Evaluation of LLMs} Our experiments were constrained to a single run of GPT-3.5, GPT-4, and LLaMa, which we aim to represent the state-of-the-art LLMs. This limitation is due to the prohibitive cost of running more extensive evaluations across a broader range of popular LLMs, potentially omitting insights from newer or less known models.

\section*{Ethics Statement}
To avoid potential copyright issues associated with using narratives, we only disclose the name of the original text-based game our work is based on, rather than the content of the original narratives. However, we will release the LLM-generated content based on the character relationships in original text-based game and the annotated character relations for the research community.
Please be advised that the detective narrative dataset may contain descriptions of violent events, actions, or characters. This content is included solely for academic, research, and narrative analysis purposes. It is not meant to glorify or trivialise violence in any form. Users should be aware of potential exposure to violent content and engage with the dataset professionally and responsibly. This dataset is unsuitable for minors and those sensitive to such content. 

\section*{Acknowledgements}
This work was supported in part by the UK Engineering and Physical Sciences Research Council (EPSRC) through a New Horizons grant (grant no. EP/X019063/1) and a Turing AI Fellowship (grant no. EP/V020579/2).  

\bibliography{anthology,custom}
\bibliographystyle{acl_natbib}

\appendix
\section{Data Selection}
\label{sec:data-selection}
\paragraph{Filtering}
We excluded low quality narratives from our analysis due to various reasons: these included stories that were too simple with little content to analyse, those with entirely public background information offering limited merging opportunities, and narratives featuring supernatural elements. Additionally, we omitted stories with weak character relationships, such as groups of strangers in survival or mystery settings, and those with overly complex rules involving characters with multiple abilities. Poorly formatted narratives, those with an excessive number of false identities, and settings like psychiatric hospitals where characters have compromised memories were also filtered out. 

\paragraph{Processing}
Extracting text from scanned narratives presents challenges with the incorrect recognition of text, a common issue with Chinese characters. This complexity is compounded by the presence of mixed or nested images within the content and the use of colored backgrounds. To address these challenges, we initially employed Adobe for the primary text extraction. Subsequently, we leveraged the capabilities of the ChatGPT model \texttt{GPT-3.5}, to rewrite and refine the text. This two-step process helps in correcting errors originating from the original scanned documents, ensuring a more accurate and reliable text output.

\section{Annotation Guideline}
\label{appendix:annotation_guideline}
In this section, we provide the annotation guidelines given to the annotators. For the sake of demonstration, we introduce 5 hypothetical characters: \texttt{Alpha, Bravo, Charlie, Delta}, and \texttt{Echo}. 
\subsection{Disclaimers of Risks}
Consider that the detective narrative dataset may contain descriptions of violent events, actions, or characters, we warn annotators of such content before the annotation process:
\begin{lstlisting}[mathescape=true]
Please be advised that the detective narrative dataset may contain descriptions of violent events, actions, or characters. This content is included solely for academic, research, and narrative analysis purposes. It is not meant to glorify or trivialise violence in any form. Annotators should be aware of potential exposure to violent content and engage with the dataset professionally and responsibly.
\end{lstlisting}

\subsection{Definition of Terminologies}
Before delving into the specific tasks of this project, it's crucial to define some key terms to ensure clarity and consistency:

\paragraph{Character} This term denotes any intelligent entity within the narrative context. Examples of characters include humans, animals with demonstrated intelligence (like dogs), and artificial entities such as intelligent robots. However, it's important to note that not every entity qualifies as a \textit{character}. For instance, collective groups (e.g., a gang), inanimate objects (e.g., an ordinary tree), or locations (e.g., a mansion) do not fall under the definition of a \textit{character} for this project.

\paragraph{Relationship} In the context of this project, a relationship refers to a social connection or interaction between two characters. A key aspect of relationships in this study is their directional nature, meaning that the perception of the relationship can vary depending on the character's viewpoint. For example, consider two characters, \texttt{Alpha} and \texttt{Bravo}. If they are friends, but \texttt{Bravo} is secretly aware of \texttt{Alpha}'s betrayal, then \texttt{Alpha} might still perceive \texttt{Bravo} as a friend, while \texttt{Bravo} may no longer see \texttt{Alpha} in the same light. 

\paragraph{Conflicts} in the relationship graph can arise from two main sources: (1) Character Disagreement and (2) False Relationship, both resulting from partially observed or interpreted information. Below are explanations and examples for each type: 
\textit{Character Disagreement} happens when a character deliberately attempts to deceive others by assuming the identity of someone else. For example, if a male character \texttt{Alpha} commits a murder while disguised in female character \texttt{Bravo}'s clothes and wearing a wig, witnesses might mistakenly believe that \texttt{Bravo} is the murderer.
\textit{False Relationship} can arise under various scenarios, with extramarital affairs being a common cause. In such a scenario, \texttt{Alpha} may believe that \texttt{Bravo} is their father, while in reality, their biological father is \texttt{Charlie}.

\subsection{Annotation Tasks}
In this project, you will work with multiple detective narratives, each offering a different character's perspective on the same story. For instance, if \texttt{Alpha, Bravo, Charlie,} and \texttt{Delta} are all involved in a murder case and are suspects, you will receive four separate narratives. Each narrative will present the story from the viewpoint of one of these characters. Your task is to analyse these narratives and identify the underlying relationships between characters based on the varying perspectives provided. For each annotation task, we apply the following prompts for both human annotators and baselines.

\paragraph{Character Extraction} We highlighted the variable(s) in blue in our designed prompt template below: 

\begin{lstlisting}[mathescape=true]
Identify and list all characters mentioned in the background story of (*@\myblue{\{c\}}@*). 
Narrative Background: (*@\myblue{\{Background\}}@*) 
Please provide the response in JSON format for clarity and structure, listing each character distinctly. 
Example format: {(*@``@*)characters(*@''@*): [(*@``@*)character 1(*@''@*), (*@``@*)character 2(*@''@*), ...]}
\end{lstlisting}
$c$ is the character name.

\paragraph{Entity Linking} 
Extracting inter-character relationships and outputting the relationship graph, including characters and their relationships. This prompt is used for the AllTogether and human annotation.

\begin{lstlisting}[mathescape=true]
Please classify the relationships between the characters using the given relationship categories (*@\myblue{\{categories\}}@*). Note that multiple relationships categories can be applied to each character pair.

Character Background of (*@\myblue{\{c\}}@*): (*@\myblue{\{Background\}}@*) 

Provide the relationships in the following JSON format ONLY:
{
  (*@``@*)character name": [
    [(*@``@*)linked character(*@''@*), (*@``@*)relationship(*@''@*)],
    [(*@``@*)linked character(*@''@*), (*@``@*)relationship(*@''@*)]
  ],
  (*@``@*)character name": [
    [(*@``@*)linked character(*@''@*), (*@``@*)relationship(*@''@*)],
    [(*@``@*)linked character(*@''@*), (*@``@*)relationship(*@''@*)]
  ]
}
\end{lstlisting}

Given the character list, extracting inter-character relationships and outputting the relationship graph. This prompt is used for the DirRelation.

\begin{lstlisting}[mathescape=true]
Using the given relationship categories (*@\myblue{\{categories\}}@*), please classify the relationships between the characters listed below. Note that multiple relationships categories can be applied to each character pair.

List of Characters: (*@\myblue{\{character list\}}@*)

Character Background of (*@\myblue{\{c\}}@*): (*@\myblue{\{Background\}}@*) 

Provide the relationships in the following JSON format ONLY:
{
  (*@``@*)character name": [
    [(*@``@*)linked character(*@''@*), (*@``@*)relationship(*@''@*)],
    [(*@``@*)linked character(*@''@*), (*@``@*)relationship(*@''@*)]
  ],
  (*@``@*)character name": [
    [(*@``@*)linked character(*@''@*), (*@``@*)relationship(*@''@*)],
    [(*@``@*)linked character(*@''@*), (*@``@*)relationship(*@''@*)]
  ]
}
\end{lstlisting}

Given a pair of characters, extract their relationships from the provided narrative. This prompt is used for the PairRelation.

\begin{lstlisting}[mathescape=true]
Using the given relationship categories (*@\myblue{\{categories\}}@*), classify the relationship between characters (*@\myblue{\{a\}}@*) and (*@\myblue{\{b\}}@*) based on the character background of (*@\myblue{\{c\}}@*). You may select multiple relationships. 

Character Background: (*@\myblue{\{Background\}}@*) 

Response in JSON format ONLY: {"(*@\myblue{\{a\}}@*)": ["(*@\myblue{\{b\}}@*)", "relationship 1, relationship 2"], "(*@\myblue{\{b\}}@*)": ["(*@\myblue{\{a\}}@*)", "relationship 1, relationship 2"]}
\end{lstlisting}

\subsection{Special Cases}
During our trial run, we found there are some cases we need to clarify where annotators have different opinions on it:
 \paragraph{Perspective of Annotation} When reading through detective stories, comprehend the stories from an objective point of view. For instance, if \texttt{Alpha} helped \texttt{Bravo} but \texttt{Bravo} does not appreciate the help, you shall disregard \texttt{Bravo}'s emotion and still annotate that "\texttt{Alpha} is the helper of \texttt{Bravo}".
\paragraph{Naming Convention} In our dataset, the same story will be conveyed from the perspective of numerous characters. Therefore, each character might use different names to refer to the same underlying character. For instance, if \texttt{Charles} is the son of \texttt{Alpha}, then the narrative from \texttt{Alpha}'s perspective may refer to \texttt{Charles} exclusively using a nickname, \texttt{Charlie}. In this case, the relationship graph of \texttt{Alpha} shall \emph{use the name that appeared in the narrative as the annotation}, which is \texttt{Charlie} in this case. However, when forming the overall relationship graph, \emph{always use the formal name of the character as annotation}.

\section{Relationship Schemes}
\label{appendix:relationship_schemes}
In this section, we list our relationship categories.

\begin{itemize}
    \item \textbf{Romantic Relationships}, which focus on the emotional, psychological, and sometimes physical intimacy between characters who are involved in a romantic or amorous setting. This can range from courtship to marriage and can include a variety of romantic scenarios, conflicts, or issues. 
        \begin{itemize}
             \item Wife of X
                \begin{itemize}
                 \item Concubine of X
                \end{itemize}
             \item Husband of X
             \item Extramarital Affair with X
                \begin{itemize}
                 \item Secret Lover of X
                \end{itemize}
            \item Romantic relationships with X
                \begin{itemize}
                 \item Lover of X
                 \item Boyfriend of X
                 \item Girlfriend of X
                \end{itemize}
            \item Ex-romantic relationships with X 
                \begin{itemize}
                 \item Ex-boyfriend of X
                 \item Ex-girlfriend of X
                 \item Ex-wife of X
                 \item Ex-husband of X
                \end{itemize}
            \item Admirer of X 
                \begin{itemize}
                 \item Secret Admirer of X
                 \item Fondness of X
                \end{itemize}
            \item Admired by X 
                \begin{itemize}
                 \item Liked by X
                 \item ecret Crush of X
                \end{itemize}
            \item Fiance of X
            \item Fiancee of X
            \item Co-wives of X
         \end{itemize}
    \item \textbf{Family Relationships}, which are between members of a family unit, which are often critical to the protagonist's identity or quest. This can include parent-child dynamics, sibling bonds, and other familial relationships (Grandparent, Cousin, etc.). It often involves loyalty, tradition, unconditional love, dysfunction, generational conflict, and personal growth. 
        \begin{itemize}
             \item Father of X
                \begin{itemize}
                 \item Father-in-law of X
                 \item Future Father-in-law of X
                 \item Adoptive Father of X
                 \item Step-father of X
                 \item Biological Father of X
                \end{itemize}
             \item Mother of X
                 \item Mother-in-law of X
                 \item Future Mother-in-law of X
                 \item Adoptive Mother of X
                 \item Step-Mother of X
                 \item Biological Mother of X
             \item Child of X
                \begin{itemize}
                 \item Son of X
                 \item Son-in-law of X
                 \item Future Son-in-law of X
                 \item Adoptive Son of X
                 \item Step-Son of X
                 \item Biological Son of X
                 \item Daughter of X
                 \item Daughter-in-law of X
                 \item Future Daughter-in-law of X
                 \item Adoptive Daughter of X
                 \item Step-Daughter of X
                 \item Biological Daughter of X
                \end{itemize}
            \item Sibling of X
                \begin{itemize}
                 \item Brother of X
                 \item Half Brother of X
                 \item Adoptive Brother of X
                 \item Step-brother of X
                 \item Elder Brother of X
                 \item Younger Brother of X
                 \item Twin Brother of X
                 \item Sister of X
                 \item Half Sister of X
                 \item Adoptive Sister of X
                 \item Step-sister of X
                 \item Elder Sister of X
                 \item Younger Sister of X
                 \item Twin Sister of X
                \end{itemize}
            \item Grandparent of X
                \begin{itemize}
                 \item Grandfather of X
                 \item Grandmother of X
                \end{itemize}
            \item Grandchild of X
                \begin{itemize}
                 \item Grandson of X
                 \item Granddaughter of X
                \end{itemize}
            \item Relative of X
                \begin{itemize}
                 \item Sister-in-law of X
                 \item Brother-in-law of X
                 \item Nephew of X
                 \item Aunt of X
                 \item Uncle of X
                 \item Niece of X
                 \item Cousin of X
                 \item Future Relative of X
                \end{itemize}
            \item Possibly Family of X
         \end{itemize}
    \item \textbf{Ally Relationships}, which involve a cooperative bond formed between characters, often to achieve a common goal or to face a shared enemy. This relationship is usually strategic and can be either temporary or long-term. These relationships may not be rooted in emotional bonds but rather strategic interests. 
        \begin{itemize}
            \item Friend of X
                \begin{itemize}
                 \item Sworn Brother of X
                \end{itemize}
            \item Mentor of X
                \begin{itemize}
                 \item Teacher of X
                \end{itemize}
            \item Student of X
            \item Informant of X
            \item Information Receiver from X
            \item Acquaintance of X
                \begin{itemize}
                 \item Classmate of X
                 \item Schoolmate of X
                 \item Neighbour of X
                \end{itemize}
            \item Helper of X
                \begin{itemize}
                 \item Saviour of X
                \end{itemize}
            \item Helped by X
                \begin{itemize}
                 \item Seeker of Help from X
                 \item Saved by X
                \end{itemize}
            \item Quest Companion of X
                \begin{itemize}
                 \item Crime Partner of X
                \end{itemize}
            \item Guest of X
            \item Host of X
        \end{itemize}
    \item \textbf{Antagonistic Relationships} refers to connections between individuals or groups at odds due to conflicting beliefs, loyalties, or goals. 
        \begin{itemize}
            \item Perpetrator of X
                \begin{itemize}
                 \item Bully of X
                 \item Murderer of X
                \end{itemize}
            \item Attempted Perpetrator of X
                \begin{itemize}
                 \item Attempted Murderer of X
                \end{itemize}
            \item Victim of X
                \begin{itemize}
                 \item Killed by X
                \end{itemize}
            \item Dislike of X
                \begin{itemize}
                 \item Jealous of X
                 \item Unsuccessful Helper of X
                 \item Suspect to X
                \end{itemize}
            \item Disliked by X
                \begin{itemize}
                 \item Jealous by X
                \end{itemize}
            \item Subject of investigation for X
            \item Deceiver of X
                \begin{itemize}
                 \item Thief of X
                 \item Betrayer of X
                 \item Spy of X
                \end{itemize}
            \item Deceived by X
                \begin{itemize}
                 \item Betrayed by X
                \end{itemize}
            \item Suspected by X
                \begin{itemize}
                 \item Suspicious of X
                 \item Suspect of X
                \end{itemize}
            \item Adversary of X
                \begin{itemize}
                 \item Enemy of X
                 \item Hate of X
                 \item Hated by X
                 \item Rebellion against X
                 \item Rival of X
                 \item Rival in Love of X
                 \item X's Victim's Family
                 \item X's Enemy's Family
                 \item Perpetrator of X's Family
                 \item in the Lawsuit against X
                \end{itemize}
            \item Manipulator of X
            \item Manipulated by X
        \end{itemize}
    \item \textbf{Business Relationships}, refer to interactions and connections between characters that are primarily based on commercial, financial, or professional contexts. These relationships are often characterized by mutual interests, professional collaborations, or economic transactions. 
        \begin{itemize}
            \item Superior of X
                \begin{itemize}
                 \item Authority over X
                 \item Employer of X
                 \item Master of X
                \end{itemize}
            \item Subordinate of X
                \begin{itemize}
                 \item Employee of X
                 \item Servant of X
                 \item Guard of X
                 \item Tour Guide of X
                 \item Minion of X
                \end{itemize}
            \item Debtor of X
            \item Creditor of X
            \item Colleague of X
            \item Business Partner of X
            \item Customer of X
                \begin{itemize}
                 \item Buyer of X
                 \item Tenant of X
                 \item Patient of X
                \end{itemize}
            \item Product Provider of X
                \begin{itemize}
                 \item Seller of X
                 \item Landlord of X
                \end{itemize}
            \item Service Provider of X
                \begin{itemize}
                 \item Lawyer of X
                 \item Messenger of X
                 \item Doctor of X
                \end{itemize}
        \end{itemize}
    \item \textbf{Special Relationships} In addition to the standard relationship labels, we also introduce a set of special relationship labels.
        \begin{itemize}
            \item Same Person as X (Different Reference)
            \item Same Person as X (Different Identity)
            \item Replaced X's Identity
            \item Stranger to X
        \end{itemize}
    ``Same Person as X (Different Reference)'' refers to different references for the same person, such as ``Costa'', ``Head Nurse Costa'', and ``Sylvia Costa''. These can easily be identified by humans as referring to the same individual. ``Same Person as X (Different Identity)'' refers to a single individual having multiple identities. For example, ``Elaine Carter'' may be known as a doctor in a hospital, but in reality, she is a criminal detective named ``Yilin Carter''. ``Elaine Carter'' is her false identity used to investigate an organ trafficking case in the hospital.

    ``Stranger'' is primarily designed for PairRelation. In human annotation, we simply skip relationships not described in the narrative. However, with PairRelation, we might ask the model about the relationships between two characters who do not know each other. Therefore, we have designed this category for such situations and remove all relationship categories labelled as ``Stranger to X'' during the evaluation stage.
\end{itemize}

\subsection{Inter-annotator Agreement}
Calculation of the F1-score for inter-annotator agreement follows the previous works \citep{semeval2007-girju, drug-effects2012, aste-li2023}.
For example, if the characters annotated by annotator 1 is $C_1$, the characters annotated by annotator 2 is $C_2$, then we have precision = $\frac{|C_1 \cap C_2|}{|C_1|}$, recall = $\frac{|C_1 \cap C_2|}{|C_2|}$, and the F1-score = $\frac{|C_1 \cap C_2|^2}{|C_1||C_2|}$.

\begin{table*}[ht!]
\centering
\resizebox{0.75\linewidth}{!}{
\begin{tabular}{cccccccccc}
\toprule
\multirow{2}{*}{Strategy} & \multicolumn{3}{c}{Llama2} & \multicolumn{3}{c}{GPT-3.5} & \multicolumn{3}{c}{GPT-4} \\ \cmidrule(lr){2-4} \cmidrule(lr){5-7} \cmidrule(lr){8-10}
  & precision      & recall     & f1     & precision     & recall    & f1    & precision    & recall    & f1   \\ 
\midrule
\multicolumn{10}{c}{\textit{Information from Single Character’s Perspective (detailed-level)}}  \\ \midrule
AllTogether & 0.096 & 0.040 & 0.057 & 0.043 & 0.031 & 0.036 & 0.202 & 0.192 & 0.197 \\ 
DirRelation & 0.116 & 0.062 & 0.081 & 0.078 & 0.050 & 0.061 & 0.149 & 0.182 & 0.164 \\ 
PairRelation & 0.017 & 0.061 & 0.027 & 0.018 & 0.061 & 0.028 & 0.032 & 0.173 & 0.054 \\ 
\midrule
\multicolumn{10}{c}{\textit{Information from All Characters’ Perspectives (detailed-level)}}  \\ \midrule
AllTogether & 0.097 & 0.016 & 0.028 & 0.049 & 0.016 & 0.024 & 0.188 & 0.058 & 0.088 \\ 
DirRelation & 0.153 & 0.025 & 0.043 & 0.049 & 0.020 & 0.028 & 0.131 & 0.049 & 0.071 \\ 
PairRelation & 0.008 & 0.065 & 0.015 & 0.009 & 0.083 & 0.016 & 0.010 & 0.165 & 0.019 \\ 
\midrule
\multicolumn{10}{c}{\textit{Information from Single Character’s Perspective (medium-level)}}  \\ \midrule
AllTogether & 0.129 & 0.054 & 0.076 & 0.056 & 0.041 & 0.047 & 0.283 & 0.269 & 0.276 \\ 
DirRelation & 0.160 & 0.085 & 0.111 & 0.119 & 0.076 & 0.093 & 0.219 & 0.267 & 0.240 \\ 
PairRelation & 0.025 & 0.089 & 0.039 & 0.029 & 0.099 & 0.045 & 0.047 & 0.258 & 0.080 \\ 
\midrule
\multicolumn{10}{c}{\textit{Information from All Characters’ Perspectives (medium-level)}}  \\ \midrule
AllTogether & 0.121 & 0.02 & 0.035 & 0.064 & 0.020 & 0.031 & 0.267 & 0.082 & 0.125 \\ 
DirRelation & 0.203 & 0.033 & 0.057 & 0.092 & 0.037 & 0.053 & 0.202 & 0.075 & 0.110 \\ 
PairRelation & 0.012 & 0.092 & 0.021 & 0.014 & 0.132 & 0.025 & 0.014 & 0.238 & 0.027 \\ 
\midrule
\multicolumn{10}{c}{\textit{Information from Single Character’s Perspective (high-level)}}  \\ \midrule
AllTogether & 0.205 & 0.085 & 0.120 & 0.094 & 0.069 & 0.080 & 0.417 & 0.397 & 0.407 \\ 
DirRelation & 0.256 & 0.137 & 0.178 & 0.236 & 0.151 & 0.184 & 0.320 & 0.391 & 0.352 \\ 
PairRelation & 0.054 & 0.190 & 0.084 & 0.079 & 0.268 & 0.122 & 0.093 & 0.505 & 0.157 \\ 
\midrule
\multicolumn{10}{c}{\textit{Information from All Characters’ Perspectives (high-level)}}  \\ \midrule
AllTogether & 0.223 & 0.038 & 0.064 & 0.089 & 0.029 & 0.043 & 0.450 & 0.138 & 0.211 \\ 
DirRelation & 0.342 & 0.056 & 0.096 & 0.167 & 0.067 & 0.096 & 0.334 & 0.124 & 0.181 \\ 
PairRelation & 0.030 & 0.236 & 0.054 & 0.041 & 0.383 & 0.073 & 0.032 & 0.524 & 0.060 \\ 
\bottomrule
\end{tabular}}
\caption{Evaluation of three different relation detection strategies to extract relationships with Chinese narrative input.} 
\label{tab:evaluation-f1-all-category}
\end{table*}

\section{Evaluation} \label{sec:evaluation-smilarity}
\subsection{Performance Across Different Category Levels}

Table~\ref{tab:evaluation-f1-all-category} presents the performance of three LLMs – Llama2, GPT-3.5, and GPT-4 – across different relation detection strategies and levels of relationship categorisation. The performance is measured in terms of precision, recall, and F1-score. The relation detection strategies tested are AllTogether, DirRelation, and PairRelation, and the relationship categorisation levels are detailed, medium, and high. The analysis is divided based on whether the information was derived from a single character's perspective or from all characters' perspectives.

As previously mentioned, we developed a hierarchical structure for relationship categories, consisting of 5 top-level categories, 54 intermediate categories, and 163 detailed categories. In our main text, due to page limitations, we report only the results for the intermediate categories, which we consider the most representative. Detailed categories may have some degree of overlap, while top-level categories are too broad and may lose essential information. We present the full results across all category levels, affirming the observations consistent with those reported in the main text. 

\paragraph{Strategy Performance Variability}
AllTogether: Generally, GPT-4 performs better in this strategy across all levels, indicating its superior ability to extract and understand relationships without additional character inputs.
DirRelation: This strategy shows a consistent pattern of improved performance across all models, especially at the high-level categorization. This suggests that DirRelation, which involves direct relationship extraction, might be a more suitable approach for these models.
PairRelation: Generally, this strategy has the lowest F1-scores among the three, which might be due to its nature of generating more hallucinations as it prompts for relationships between every pair of characters. We also realised that low precision, resulting from such fabrications, had a more detrimental effect than low recall, as it could lead to complete misinformation for readers. For example, while failing to detect that ``Drake'' is the secret lover of ``Gale'' is a shortcoming, the model incorrectly predicting ``Drake'' as the father of ``Gale'' is a more severe error.

\paragraph{Impact of Perspective}
The models generally perform better when extracting information from a single character’s perspective compared to all characters’ perspectives. This could be due to the increased complexity and potential for contradictory information when considering multiple perspectives.

\paragraph{Category Level Impact}
Detailed-level: The models show relatively lower scores, likely due to the finer granularity of relationships, which increases complexity.
Medium-level: This level appears to strike a balance between detail and breadth, with GPT-4 showing notably higher scores in the AllTogether strategy.
High-level: Here, the models generally achieve higher F1-scores, particularly GPT-4, suggesting that broader categories are easier for the models to identify and classify.

\paragraph{Model Comparison}
GPT-4: This model consistently outperforms the others in almost all scenarios, especially in the AllTogether strategy, indicating its advanced capability in understanding complex relationships.
Llama2 and GPT-3.5: These models show improvements in the DirRelation strategy, particularly at higher-level categories, but fall short compared to GPT-4.

\begin{table*}[ht!]
\centering
\resizebox{0.75\linewidth}{!}{
\begin{tabular}{cccccccccc}
\toprule
\multirow{2}{*}{Strategy} & \multicolumn{3}{c}{Llama2} & \multicolumn{3}{c}{GPT-3.5} & \multicolumn{3}{c}{GPT-4} \\ \cmidrule(lr){2-4} \cmidrule(lr){5-7} \cmidrule(lr){8-10}
  & Pre      & Rec     & F1     & Pre     & Rec    & F1    & Pre    & Rec    & F1   \\ \midrule
\multicolumn{10}{c}{\textit{Information from Single Character’s Perspective}}  \\ \midrule
DirRelation(Extracted)     & 0.160 & 0.085 & 0.111 & 0.119 & 0.076 & 0.093 & 0.219 & 0.267 & 0.240 \\
DirRelation(Given)     & 0.170 & 0.097 & 0.124 & 0.155 & 0.102 & 0.123 & 0.293 & 0.342 & 0.315 \\
PairRelation(Extracted)    & 0.025 & 0.089 & 0.039 & 0.029 & 0.099 & 0.045 & 0.047 & 0.258 & 0.080 \\
PairRelation(Given)    & 0.045 & 0.181 & 0.071 & 0.056 & 0.158 & 0.083 & 0.105 & 0.402 & 0.167 \\
\midrule
\multicolumn{10}{c}{\textit{Information from All Characters’ Perspectives}}  \\ \midrule
DirRelation(Extracted)      & 0.203 & 0.033 & 0.057 & 0.092 & 0.037 & 0.053 & 0.202 & 0.075 & 0.110 \\
DirRelation(Given)      & 0.163 & 0.032 & 0.053 & 0.155 & 0.040 & 0.064 & 0.330 & 0.115 & 0.171 \\
PairRelation(Extracted)     & 0.012 & 0.092 & 0.021 & 0.014 & 0.132 & 0.025 & 0.014 & 0.238 & 0.027 \\
PairRelation(Given)     & 0.027 & 0.168 & 0.046 & 0.034 & 0.147 & 0.055 & 0.071 & 0.282 & 0.113 \\

\bottomrule
\end{tabular}}
\caption{Detailed Comparison of Character Extraction Impact.} 
\label{tab:evaluation-f1-single-details}
\end{table*}

\subsection{Translated Version} \label{sec:translated version}

For our study, we primarily report results using the Chinese version for two main reasons. Firstly, the source material is originally in Chinese, so utilising the native language minimizes information loss during translation. Secondly, our experiments partly rely on the OpenAI API, which imposes content restrictions. Detective narratives often involve themes like murder, which can trigger OpenAI's stringent content filters, especially in English. However, these restrictions are less frequent with Chinese inputs, likely due to biases in the training data of the LLMs.

We initially attempted to translate the Chinese version into English using GPT-3.5 and even GPT-4. However, we encountered challenges in controlling the translation process with these LLMs. Despite specifying in our prompts that we required direct translation, the LLMs sometimes performed summarisation instead. Additionally, the translations were prone to hallucinations, introducing inaccuracies. Consequently, we decided to use Google Translator for the translation process to ensure more accurate and direct translations.

During the process of translating Chinese names into English, we observed that Google Translator inconsistently applied phonetic transliteration and semantic translation. This inconsistency resulted in the translated character names being quite confusing, as the approach to translation varied unpredictably between literal phonetic equivalents and meaning-based interpretations. 
Therefore, we initially used GPT-4 to translate all labelled character names from Chinese into English. Once the character names were translated, we replaced the original names in the corpus with these English versions. Following this step, we employed Google Translator to translate the entire corpus. This two-step process ensured a more consistent and clear translation of character names, while also effectively translating the broader narrative content.

In addition to our previous findings, we also report on experiments conducted using the English corpus. These experiments corroborate the results obtained from the Chinese corpus. However, it's important to note that we limited these experiments to running on Llama2 and GPT-3.5 models. The decision to exclude GPT-4 from these tests was due to its higher costs.

\begin{table}[ht!]
\centering
\resizebox{0.7\linewidth}{!}{
\begin{tabular}{ccccccc}
\toprule
\multirow{2}{*}{Strategy}                    &       \multicolumn{2}{c}{Llama2} & \multicolumn{2}{c}{GPT-3.5}  \\ \cmidrule(lr){2-3} \cmidrule(lr){4-5} 
& before & after & before & after \\ \midrule
AllTogether         & 0.312  &  \textbf{0.282}  &   0.426  &  \textbf{0.107}   \\
DirRelation          & 0.245  &  \textbf{0.219}  &  0.296  &  \textbf{0.063}      \\
PairRelation         &  0.226 &  \textbf{0.185}  &   0.303  & \textbf{0.105}    \\
\bottomrule
\end{tabular} }
\caption{Corruption Rate for English Corpus. The percentage of output relationships that fail to comply with the provided format guidelines before and after self-correction.} 
\label{tab:evaluation-corruption-rate-english}                       
\end{table}

The data indicates that self-correction significantly reduces the corruption rate across all strategies and models, with particularly notable improvements observed in GPT-3.5. GPT-3.5 benefits more from self-correction compared to Llama2, as evidenced by the greater percentage reductions in corruption rates. While all strategies show improvement with self-correction, the AllTogether strategy displays the most significant improvement for GPT-3.5, suggesting that this strategies might be more prone to initial corruption but also benefits greatly from correction.

\begin{table}[ht!]
\centering
\resizebox{0.7\linewidth}{!}{
\begin{tabular}{ccccc}
\toprule
\multirow{2}{*}{Strategy}                    &       \multicolumn{2}{c}{Llama2} & \multicolumn{2}{c}{GPT-3.5}  \\ \cmidrule(lr){2-3} \cmidrule(lr){4-5} 
& before & after & before & after \\ \midrule
AllTogether         & 0.069  &  \textbf{0.070}  &  0.044  &  \textbf{0.060}    \\
DirRelation          &  0.088 &  \textbf{0.090}  &  0.038  &  \textbf{0.043}    \\
PairRelation         & 0.025  &  0.025  &  0.014  & 0.014    \\
\bottomrule
\end{tabular} }
\caption{Performance Comparison for English Corpus. The F1-score before and after self-correction.} 
\label{tab:evaluation-f1-after-self-correction-english}                       
\end{table}

The AllTogether and DirRelation show some improvement with self-correction. However, the improvements for Llama2 are generally minimal, and the PairRelation strategy does not benefit from self-correction. This suggests that while self-correction can lead to improvements in model performance, the extent of these improvements varies and can be limited for certain strategies or models.

\begin{table}[ht!]
\centering
\resizebox{\linewidth}{!}{
\begin{tabular}{ccccccc}
\toprule
\multirow{2}{*}{Strategy} & \multicolumn{3}{c}{Llama2} & \multicolumn{3}{c}{GPT-3.5}  \\ \cmidrule(lr){2-4} \cmidrule(lr){5-7} 
  & precision      & recall     & f1     & precision     & recall    & f1       \\ \midrule
\multicolumn{7}{c}{\textit{Information from Single Character’s Perspective }}  \\ \midrule
DirCharacter  &  0.628 & 0.634 & 0.631 & 0.633 & 0.631 & 0.632  \\
AllTogether & 0.808 & 0.528 & 0.639  &  0.522 & 0.572 & 0.546 \\
DirRelation & 0.718 & 0.619 & 0.665 &   0.663 & 0.620 & 0.641 \\
PairRelation &  0.692 & 0.645 & 0.668 &  0.575 & 0.637 & 0.605 \\
\midrule
\multicolumn{7}{c}{\textit{Information from All Characters’ Perspectives}}  \\ \midrule
DirCharacter  &  0.450 & 0.744 & 0.561 &  0.485 & 0.776 & 0.597\\
AllTogether &   0.840 & 0.310 & 0.453  & 0.776 & 0.305 & 0.438\\
DirRelation &  0.789 & 0.346 & 0.481 &  0.575 & 0.618 & 0.595 \\
PairRelation &  0.502 & 0.557 & 0.528 &   0.370 & 0.765 & 0.499 \\
\bottomrule
\end{tabular}}
\caption{F1-score of Character Extraction for English corpus.} 
\label{tab:evaluation-character-english}
\end{table}

In the context of a single character’s perspective, all strategies show relatively good performance. From all characters' perspectives, the performance of both models decreases, indicating the complexity of handling multiple viewpoints. In both perspectives, DirRelation and PairRelation show more balanced precision and recall compared to AllTogether, especially for GPT-3.5. The PairRelation, while showing reasonable performance in a single character’s perspective, drops in effectiveness when considering all characters' perspectives. 

\begin{table}[ht!]
\centering
\resizebox{\linewidth}{!}{
\begin{tabular}{ccccccc}
\toprule
\multirow{2}{*}{Strategy} & \multicolumn{3}{c}{Llama2} & \multicolumn{3}{c}{GPT-3.5}  \\ \cmidrule(lr){2-4} \cmidrule(lr){5-7} 
  & precision      & recall     & f1     & precision     & recall    & f1       \\ \midrule
\multicolumn{7}{c}{\textit{Information from Single Character’s Perspective (detailed-level)}}  \\ \midrule
AllTogether & 0.136 & 0.081 & 0.102 & 0.088 & 0.076 & 0.082 \\
DirRelation & 0.147 & 0.113 & 0.128 & 0.095 & 0.081 & 0.088 \\
PairRelation & 0.022 & 0.11 & 0.037 & 0.016 & 0.072 & 0.026 \\

\midrule
\multicolumn{7}{c}{\textit{Information from All Characters’ Perspectives (detailed-level)}}  \\ \midrule
AllTogether & 0.144 & 0.029 & 0.049 & 0.112 & 0.022 & 0.037 \\
DirRelation & 0.17 & 0.041 & 0.066 & 0.046 & 0.024 & 0.031 \\
PairRelation & 0.011 & 0.09 & 0.019 & 0.005 & 0.068 & 0.009 \\

\midrule
\multicolumn{7}{c}{\textit{Information from Single Character’s Perspective (medium-level)}}  \\ \midrule
AllTogether & 0.178 & 0.107 & 0.134 & 0.121 & 0.104 & 0.112 \\
DirRelation & 0.191 & 0.147 & 0.166 & 0.132 & 0.113 & 0.122 \\
PairRelation & 0.032 & 0.157 & 0.053 & 0.026 & 0.116 & 0.042 \\

\midrule
\multicolumn{7}{c}{\textit{Information from All Characters’ Perspectives (medium-level)}}  \\ \midrule
AllTogether & 0.208 & 0.042 & 0.07 & 0.181 & 0.036 & 0.06 \\
DirRelation & 0.232 & 0.056 & 0.09 & 0.063 & 0.032 & 0.043 \\
PairRelation & 0.014 & 0.121 & 0.025 & 0.008 & 0.1 & 0.014 \\

\midrule
\multicolumn{7}{c}{\textit{Information from Single Character’s Perspective (high-level)}}  \\ \midrule
AllTogether & 0.28 & 0.167 & 0.209 & 0.194 & 0.167 & 0.179 \\
DirRelation & 0.27 & 0.208 & 0.235 & 0.222 & 0.19 & 0.205 \\
PairRelation & 0.062 & 0.308 & 0.103 & 0.059 & 0.268 & 0.097 \\

\midrule
\multicolumn{7}{c}{\textit{Information from All Characters’ Perspectives (high-level)}}  \\ \midrule
AllTogether & 0.345 & 0.07 & 0.117 & 0.307 & 0.061 & 0.102 \\
DirRelation & 0.397 & 0.095 & 0.154 & 0.151 & 0.078 & 0.103 \\
PairRelation & 0.029 & 0.246 & 0.052 & 0.023 & 0.301 & 0.042 \\

\bottomrule
\end{tabular}}
\caption{Evaluation of three different strategies to extract relationships with English narrative input.} 
\label{tab:evaluation-f1-single-english}
\end{table}

AllTogether: Generally, this strategy shows moderate to high precision but lower recall, leading to varied F1-scores. Its performance is more consistent in the single character’s perspective.
DirRelation: Exhibits balanced precision and recall, generally performing better than AllTogether in terms of F1-scores. It shows a notable improvement in F1 scores at the high-level perspective.
PairRelation: Significantly lower precision across all categories, but higher recall in high-level information extraction. However, the low precision greatly affects its overall F1-scores.

After translation, all strategies exhibit improved performance with Llama2, which shows its better understanding ability in English. GPT was also shown to have better performance in English corpus \cite{gpt4}, however, in our experiment, GPT shows varying results with the English corpus compared to the Chinese one, likely due to content restrictions. Detective narratives often include themes like murder, which may activate OpenAI's strict content filters. However, such restrictions are less common with Chinese inputs, probably because of biases in the training data of LLMs, resulting in decreased performance with the English corpus. 

\subsection{Similarity Distance}
We attempted to incorporate cosine similarity measurement scores alongside F1-scores due to the unclear boundaries between some categories, and because LLMs sometimes produce correct outputs that don't match the specified category. For instance, \emph{``mother of x''} might be output as \emph{``mother to x''}, which would be marked incorrect in metrics like accuracy and F1-score. The similarity scores aim to mitigate the impact of such formatting errors.

However, we discovered that cosine similarity scores are not a reliable measure, as they can yield unreasonable results. For instance, the similarity score between \emph{``employer''} and \emph{``stranger''} is 0.29, but it's 0.42 between \emph{``daughter''} and \emph{``stranger''}, suggesting that a daughter is closer to being a stranger than an employer. This result is counterintuitive from a human perspective, leading us to exclude these similarity measurement results in this paper.



\end{document}

%% file: Sections/introduction_v1.tex
\section{Introduction}\label{sec:intro}

Tasks like multi-agent interaction \citep{generative-agent2023, werewolf2023, avalon2023} and character-centric narrative understanding \citep{Zhu2023AreNM} have recently gained significant attention. These tasks require a deeper understanding of complex 
relationships among multiple entities \citep{worth2004narrative}, thereby serving as a critical benchmark for assessing the reasoning capabilities of LLMs \citep{bubeck2023gpt4AGI}. Detective stories, where characters often adopt multiple identities or aliases that are revealed at various points, are the most appropriate testbed for assessing LLMs' capability of deducing complex relationships. However, existing datasets designed for character-centric narrative understanding are either built on well-known stories that LLMs have been trained on \citep{brahman-etal-2021-characters-tell, sang-etal-2022-tvshowguess, chen2023harrypotter, dynamic-fictional-relationships2016}, or consist of simpler texts \citep{bamman-etal-2020-annotated, heroes-stammbach-etal-2022, fairytaleqa-xu-etal-2022}, such as children's stories, where characters and their relationships are typically introduced when the characters first appear in the narrative \citep{narrativeplay2023}. 

\begin{figure}[t!]
    \centering
    \includegraphics[width=\linewidth]{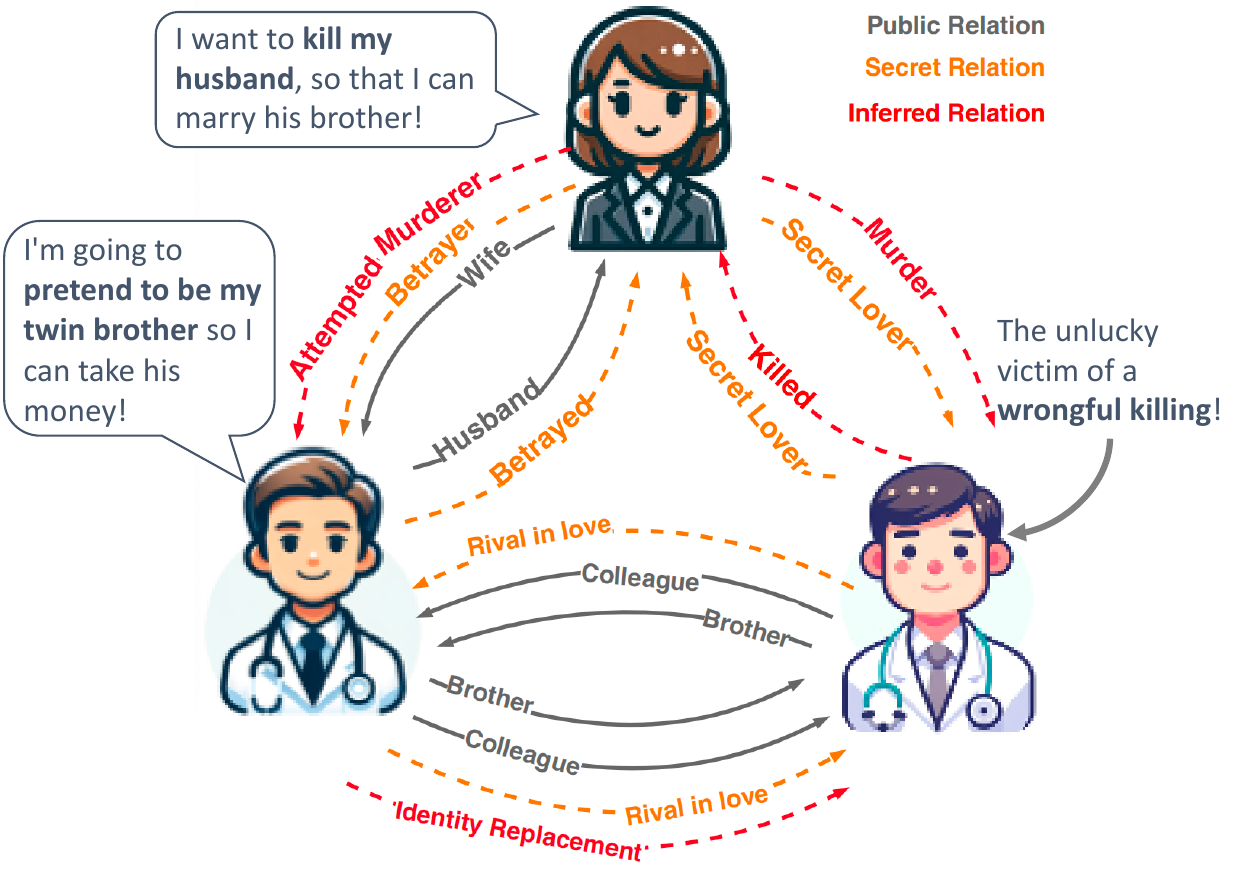}
    \caption{The example illustrates complex relationships of characters in narratives. 
    \textcolor{gray}{Gray-colored} relationships represent surface-level information, widely known to most characters. \textcolor{orange}{Orange-colored} relationships, on the other hand, are secrets known to only one or very few individuals, often conflicting with the commonly known relationships; these are referred to as secret relationships. \textcolor{red}{Red-colored} relationships represent inferred information, meaning they are not explicitly stated in any character's story but can be deduced by synthesising information from all characters collectively. LLMs struggle with such complex relationships in long narratives. }
    \label{fig:relationships}
\end{figure}

However, relationships between characters are often characterised by \textbf{incomplete and uncertain} information in reality. These references may involve descriptions from another character's perspective, as in \emph{``A middle-aged man walked out of the main gate, with grey and white hair, and a slender build.''}.
Also, in real social scenarios, this is often not the case. For instance, as depicted in Figure \ref{fig:relationships}, husband \textit{A} may remain unaware that his brother \textit{B} is the secret lover of his wife \textit{C} and the biological father of his son \textit{D}. Each individual may have differing interpretations of their relationships, with these perceptions potentially \textbf{conflicting} based on their own perspectives.

Misidentified relationships can greatly impact the core conflicts and plot of a story, as these complex relationships are often central to the narrative. Therefore, we have developed the benchmark, COntextual Narrative ANalysis (\emph{Conan})
to understand complex relationships in detective narratives. We have also outlined the desired input-output format, identified three sub-tasks within this framework, and developed hierarchical relationship categories for evaluation, drawing insights from the field of social science. 
\emph{Conan} is constructed to extract role-oriented relational graphs from detective narratives. It comprises detective narratives from various characters' perspectives, along with annotated relationships that can be deduced from the narrative. As shown in Figure~\ref{fig:relationships}, it includes three types of relationships: (1) \emph{Public Relations} that are known to most people; (2) \emph{Secret Relations} that are known only to a few or even just one person and often conflicting with the widely known relationships; (3) \emph{Inferred Relations}, which are not explicitly articulated in any single character’s story and must be deduced by combining information from multiple characters. 
This dataset can act as a benchmark to test the cognitive and inferential abilities of LLMs. Additionally, it holds the potential to improve LLM capabilities in areas such as narrative comprehension \citep{narrativeplay2023}, simulation agents \citep{generative-agent2023}, and game agents \citep{werewolf2023, avalon2023}.

Our quantitative experiments and qualitative analysis show our benchmark is challenging for cutting-edge LLMs (GPT-3.5, GPT-4, and Llama2).
Our findings reveal that these models struggle primarily due to two reasons: (1) the complexity of information that necessitates inferential reasoning; (2) the length of narratives which demand efficient key information extraction. Additionally, our evaluation across three distinct strategies has pinpointed the most effective strategy for various scenarios, while also highlighting those that are inefficient and underperforming. Given the high operational costs of LLMs, these insights are valuable for saving time and resources in future research\footnote{Our code and dataset are available at 
\url{https://github.com/BLPXSPG/Conan}}. 

In summary, we have made the following contributions:
\begin{itemize}[noitemsep,nolistsep]
    

    \item We have constructed and annotated the \emph{Conan} dataset for the task, designed to evaluate and understand the inference capacity of LLMs. 

    \item We have designed hierarchical relationship categories for evaluation, built on insights from social science and necessary empirical observations during the process of manual annotation and LLMs' evaluation.
    
    \item To assess the performance of advanced LLMs, namely, GPT-3.5, GPT-4, and Llama2, we have conducted evaluations with three different strategies using our benchmark and have identified the most effective strategy for various scenarios. 
    
    \item Our findings reveal that LLMs significantly underperform humans on \emph{Conan}. We have carried out a series of experiments to validate our hypotheses regarding the causes of LLMs' failure. 
    
\end{itemize}